\documentclass[letterpaper, 10pt, conference]{ieeeconf}  

\IEEEoverridecommandlockouts                              

\usepackage{graphicx} 
\usepackage{amsmath} 
\usepackage{amssymb}  
\usepackage{hyperref}
\usepackage{multirow}
\usepackage{booktabs}
\usepackage{xcolor}
\usepackage{tabularx}
\usepackage{caption}
\usepackage{subcaption}

\title{\LARGE \bf
Sampling-based Uncertainty Estimation\\ for an Instance Segmentation Network
}

\author{Florian Heidecker$^1$, Tobias Susetzky$^2$, Erich Fuchs$^2$, and Bernhard Sick$^1$
\thanks{$^{1}$F. Heidecker, and B. Sick are with the Intelligent Embedded Systems Lab, University of Kassel, Kassel, Germany
	    {\tt\footnotesize \{florian.heidecker $\mid$ bsick\}@uni-kassel.de,}}%
\thanks{$^{2}$T. Susetzky, and E. Fuchs are with FORWISS, University of Passau, Passau, Germany 
		{\tt\footnotesize \{susetzky $\mid$ fuchse\}@forwiss.uni-passau.de}}%
}

\author{Florian Heidecker$^1$, 
        Ahmad El-Khateeb$^1$ 
        and
        Bernhard Sick$^1$ 
    \thanks{$^{1}$Florian Heidecker, Ahmad El-Khateeb and Bernhard Sick are with Intelligent Embedded Systems, University of Kassel, Wilhelmshöher Allee 73, 34121 Kassel, Germany, {\tt\small \{florian.heidecker, akhateeb, bsick\}@uni-kassel.de}}
}

\begin{document}
    \maketitle
    \thispagestyle{empty}
    \pagestyle{empty}

    \begin{abstract}
    The examination of uncertainty in the predictions of machine learning (ML) models is receiving increasing attention. One uncertainty modeling technique used for this purpose is Monte-Carlo (MC)-Dropout, where repeated predictions are generated for a single input. Therefore, clustering is required to describe the resulting uncertainty, but only through efficient clustering is it possible to describe the uncertainty from the model attached to each object. This article uses Bayesian Gaussian Mixture (BGM) to solve this problem. In addition, we investigate different values for the dropout rate and other techniques, such as focal loss and calibration, which we integrate into the Mask-RCNN model to obtain the most accurate uncertainty approximation of each instance and showcase it graphically.
\end{abstract}


\section{Introduction}
    Artificial intelligence systems are growing daily and are being used in more and more applications. Thereby new data is constantly being recorded, with new situations being added all the time \textbf{---} so there is no such thing as a computer vision dataset covering all possible tasks such as object detection, classification, and segmentation with all kinds of conditions. Therefore, the results of object detection, classification, and instance segmentation \cite{He2016, Feng2020, Qiao2021} are often insufficient or incorrect regarding the perception of some objects or instances in real-world scenarios. To address this problem, it is essential to consider the uncertainty with which machine learning (ML) models make a prediction.
    
    Many approaches exist to modeling uncertainty \cite{Malinin2018,Choi2018,Lyzhov2020,Ayhan2018,Gal2016,Blundell2015,Liu2019,Lakshminarayanan2017}. Some approaches, such as Ensemble or Monte-Carlo (MC)-Dropout, generate multiple predictions per input. The challenge is to cluster the instances of each prediction to obtain the uncertainty of each instance. For this purpose, we use the work of \cite{Heidecker2021a} as a baseline and extend or modify various aspects to obtain a good uncertainty estimation of the instances.
    
    \cite{Heidecker2021a} use Mask-RCNN~\cite{He2016} as architecture and MC-Dropout, only estimating bounding box and class score uncertainty. Therefore, we extended the model architecture by adding MC-Dropout layers to the Region Proposal Network (RPN) and mask head. By repeating the forward passes of a single input several times, we sample multiple predictions for each instance, while each of these predictions contains bounding boxes, class information, and instance masks. To increase reliability in the inferred distribution, we added focal loss~\cite{Lin2017} and calibrated~\cite{Guo2017} the model to improve the reliability and performance, especially after extending it with MC-Dropout layers. With the spatial properties of the bounding boxes, we cluster the Instances of the repetitions as in \cite{Heidecker2021a}. However, instead of applying Gaussian Mixture Model (GMM), we replace it with Bayesian Gaussian Mixture (BGM)~\cite{Blei2006} to cluster the predictions. To gain insight into the approximated uncertainty, we showcase different graphs to visualize the uncertainty of the bounding box, class score, and instance mask.

    The remainder of this article is structured as follows: Section~\ref{sec:related} provides a brief overview of uncertainty modeling and clustering techniques used in combination with MC-Dropout. Section~\ref{sec:mask_rcnn_plus} presents all modifications of the Mask-RCNN model and gives an insight into the architecture. The model's internal dependencies, analysis, and evaluation of the added focal loss, calibration, and MC-Dropout are laid out in Section~\ref{sec:eval}. After the evaluation, we present the uncertainty visualization in Section~\ref{sec:visual}. Finally, in Section~\ref{sec:conclusion}, we conclude the article's key message.
    
\section{Related Work}\label{sec:related}
    When considering uncertainty in ML, two different types of uncertainty must be distinguished. These are aleatoric uncertainty~\cite{Huellermeier2021}, which arises from the data complexity, such as label noise, and the model uncertainty, also known as epistemic uncertainty~\cite{Huellermeier2021}. In this article, we will focus on epistemic uncertainty because we want to model and determine the uncertainty that results from model architecture and the distribution space of the model's learnable parameters.

    \subsection{Uncertainty Modeling}
        There are several approaches to modeling uncertainty. In \cite{Gawlikowski2022}, the different uncertainty modeling techniques were assigned to four basic types of uncertainty prediction in ML. These four types are Single Deterministic Networks, Bayesian Methods, Ensemble Methods, and Test-Time Data Augmentation. All methods belonging to the \emph{Single Deterministic Networks}, such as Prior Networks~\cite{Malinin2018} or Mixture Density Networks~\cite{Choi2018}, do not depend on multiple predictions per input to model uncertainty. According to \cite{Gawlikowski2022}, all other types of uncertainty prediction have this prerequisite but rely on different procedures in modeling the uncertainty. \emph{Test-Time Data Augmentation} type methods such as \cite{Lyzhov2020} or \cite{Ayhan2018} use augmentations on the input data to infer the uncertainty, but with the purpose of modeling aleatoric uncertainty and not the epistemic uncertainty. In this article, we use  MC-Dropout~\cite{Gal2016}, which is counted among the \emph{Bayesian Methods} type besides Bayes by Backprop~\cite{Blundell2015} and achieves an approximation of the model uncertainty by repeating the forward pass several times with the same input data and model. Finally, there is the \emph{Ensemble Methods} type to which Deep Ensemble~\cite{Lakshminarayanan2017} and Bayesian nonparametric Ensemble~\cite{Liu2019} belong. Compared to the previous type, these models get along with one prediction per input sample because the uncertainty modeling is achieved using several different model variations.
    
    \subsection{Clustering Predictions}
        However, with the generation of multiple predictions per input, the challenge arises to cluster related instances as well as possible. In the literature, classification affinity and spatial affinity are used for this purpose. For example, in~\cite{Yang2021}, intersection over union (IoU) is applied as a spatial affinity to the 3D objects from a bird's eye view and clustered with the class labels in a soft clustering approach. Miller et al. compare four approaches for clustering in~\cite{Miller2019}, where we summarize the two BSAS-based approaches for clarity in the following overview:
        \begin{itemize}
            \item \emph{Basic Sequential Algorithmic Scheme (BSAS)}: Primarily uses the spatial affinity and calculates the IoU between the instances. If the IoU value is greater than a threshold value and thus unable to form a new cluster, the instance is assigned to the cluster with the greatest IoU score. The use of classification affinity as an additional feature is also possible.
            \item \emph{Hungarian Method}: The Hungarian Matcher~\cite{Kuhn1955} solves the $m \times n$ assignment problem. The instances of the first predictions are taken as initial clusters. Additional clusters are formed if instances cannot be assigned or more instances than clusters have been predicted.
            \item \emph{Hierarchical Density-Based Spatial Clustering of Applications with Noise (HDBSCAN)}: HDBSCAN~\cite{McInnes2017} is an extension of DBSCAN, allows clusters of different densities and is more robust against noisy data but relies on spatial affinity.
        \end{itemize}
        In addition to these approaches, \cite{Heidecker2021a} uses a GMM to cluster the predicted instances, using only the spatial affinity of the instance bounding box. Most approaches rely on the bounding box to compute the IoU of an instance \cite{Yang2021,Miller2018,Miller2019}. In~\cite{Morrison2019}, the instance mask is used instead and combined with the BSAS method. Moreover, since Mask-RCNN is also investigated in~\cite{Morrison2019}, this approach can be compared very well with ours. We noticed that using the mask for calculating the IoU is not ideal since some instances in our experiments possess bad masks. Further background details are in section \ref{sec:eval}. Because the mask of some instances is bad, it is possible that for a single instance, multiple clusters appear containing the bad predictions. This separation leads to a distorted uncertainty analysis since the clusters have been cleaned up.

\section{Modified Model Architecture}\label{sec:mask_rcnn_plus}
    \begin{figure*}[t]
        \centering
        \includegraphics[width=0.68\textwidth]{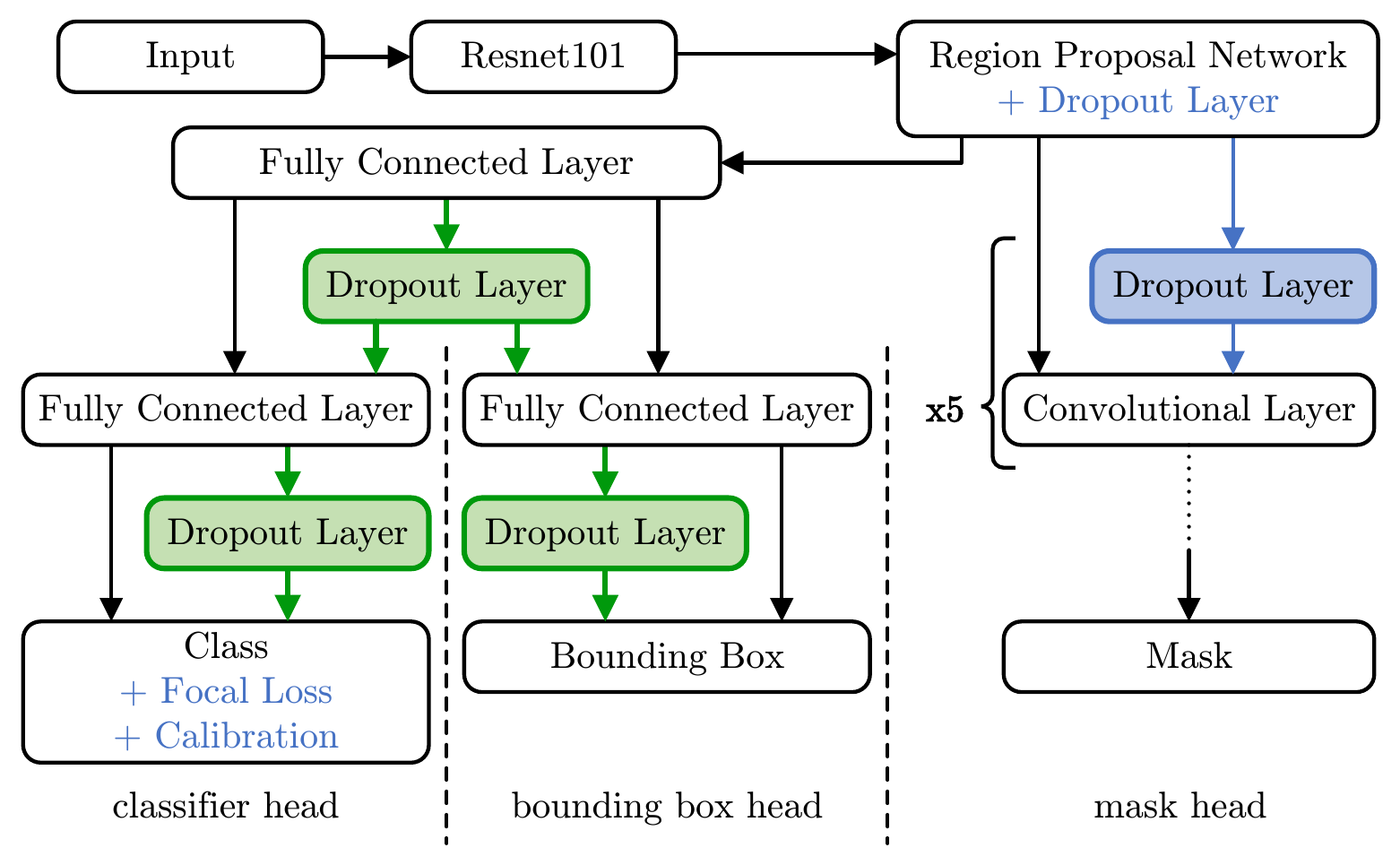}
        \caption{The image shows the extended architecture of the Mask-RCNN model~\cite{He2017}. Black represents the original Mask-RCNN model. The green boxes replaced the black arrows next to them and were introduced by \cite{Heidecker2021a}. Everything in blue indicates the model extensions in this article; the blue box also replaces the arrow next to it.}
        \label{fig:mask_rcnn_plus}
    \end{figure*}
    \begin{figure*}[t]
        \centering
        \includegraphics[width=0.68\textwidth]{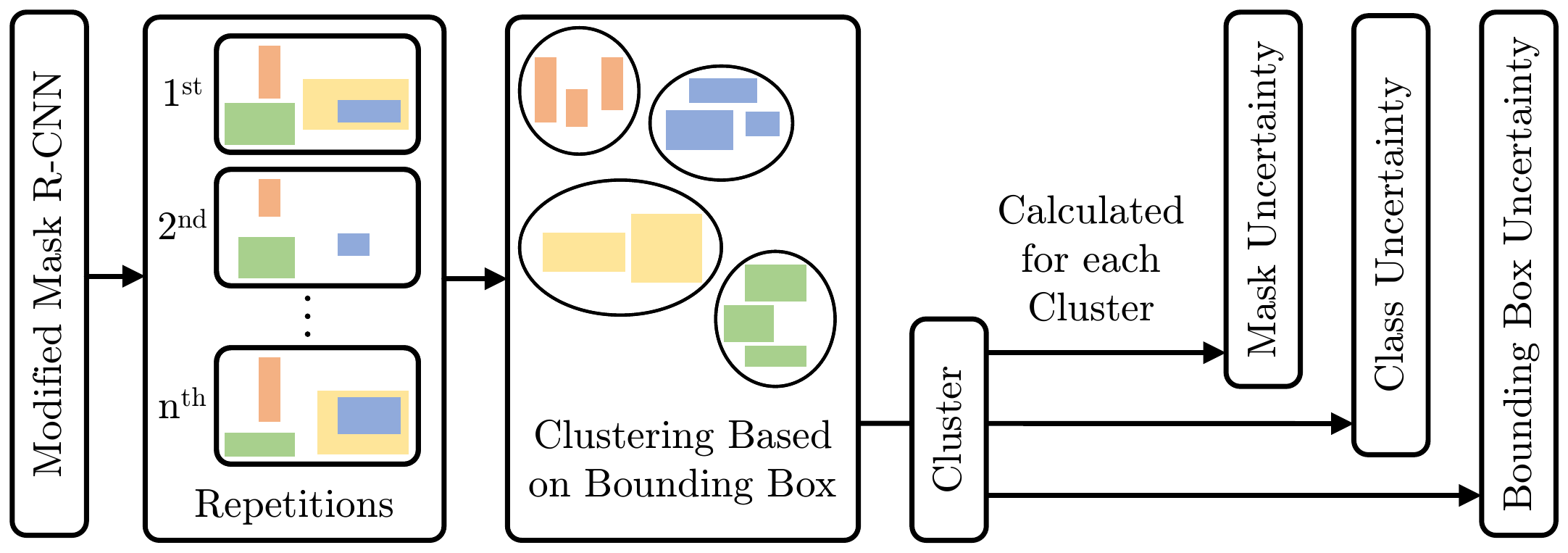}
        \caption{Uncertainty estimation via MC-Dropout followed by clustering.}
        \label{fig:cluster}
    \end{figure*}
    Mask-RCNN~\cite{He2017} is a widely used and well-known model, for instance, segmentation and provides bounding box, class, and mask proposals for each instance in an image. As a starting point, we used the PyTorch implementation of Mask-RCNN available in Torchvision~\cite{PyTorchTorchvision2022}. The architecture of Mask-RCNN is depicted in black in Fig.~\ref{fig:mask_rcnn_plus}. The backbone takes the input image and extracts features using a ResNet101~\cite{He2016}. Afterward, an RPN~\cite{Ren2017} uses the extracted features to generate region proposals. Moreover, the RPN classifies the proposals in foreground or background and ranks them according to the instance probability. The most relevant region proposals are passed through the fully connected layer in the bounding box regression head and the classification head or convolution layer in the mask head to obtain an accurate bounding box, class score, and mask for each image instance. The Mask-RCNN default model was extended in our earlier work~\cite{Heidecker2021a} by adding MC-Dropout layers~\cite{Gal2016} to the bounding box regression head and classification head to estimate the epistemic uncertainty, visualized in green in Fig.~\ref{fig:mask_rcnn_plus}. Furthermore, we modified the classification head and replaced the scalar output for the highest-scored class with a $k$-long confidence vector of $k$ classes plus the background class confidence. 
    \subsection{Mask-RCNN Postprocessing}
        Mask-RCNN has a significant difference in algorithmic procedure between training and inference~\cite{He2017}. In training mode, a region proposal is considered positive if the intersection over union (IoU) between the region proposal and the ground-truth box is higher than $0.5$ and negative otherwise. In inference mode, a defined number of region proposals with the highest score after non-maximum suppression \cite{Girshick2015} are forwarded to the heads. Because separate networks in the Mask-RCNN model perform the object detection (RPN) and the classification, the highest-scored class may be the background class for some of the top detections. Full access to the classification vector plus the background class reveals this circumstance, which means that the model is not sure that it is even an object that can be assigned with a specific class label. The original Mask-RCNN excludes the background class, erases all instances where no class score is above a user-defined threshold of $0.05$, and selects the highest-scored class to get around this issue. We have changed this slightly by only erasing those instances where the background class is above $0.45$.
    \subsection{Clustering}
    \begin{table*}[t]
        \centering
        \caption{Model performance of different Mask-RCNN architectures.}\label{tab:model_perform}
        \begin{tabular}{c|c|c|c|c|c|c}
            No. &
            Focal Loss &
            Calibration &
            \begin{tabular}[c]{@{}c@{}}MC-Dropout\\ Rate\end{tabular} & Clustering &
            \begin{tabular}[c]{@{}c@{}}Bounding Box\\ $mAP_{IoU=0.5}$\end{tabular} & 
            \begin{tabular}[c]{@{}c@{}}Mask\\ $mAP_{IoU=0.5}$\end{tabular} \\ \hline
            1  & -          & -          & -   & -   & 51,7\% & 48,8\%\\
            2  & -          & \checkmark & -   & -   & 52,2\% & 49,2\%\\ \hline
            3  & \checkmark & -          & 0.2 & BGM & 50,1\% & 48,0\%\\
            4  & \checkmark & -          & mix & BGM & 48,6\% & 46,6\%\\ \hline
            5  & \checkmark & -          & 0.5 & BGM & 48,9\% & 46,8\%\\
            6  & \checkmark & \checkmark & 0.5 & BGM & 49,0\% & 47,0\%\\
            7  & -          & -          & 0.5 & BGM & 49,1\% & 47,1\%\\
            8  & -          & \checkmark & 0.5 & BGM & 49,5\% & 47,6\%\\ \hline
            9  & \checkmark & -          & 0.5 & AGG & 46,3\% & 44,5\%\\
            10 & \checkmark & \checkmark & 0.5 & AGG & 46,9\% & 44,9\%\\
            11 & -          & -          & 0.5 & AGG & 46,8\% & 45,1\%\\
            12 & -          & \checkmark & 0.5 & AGG & 47,1\% & 45,3\%\\
        \end{tabular}
    \end{table*}
    
        As introduced in \cite{Heidecker2021a}, we repeated the forward pass of a single input $n=100$ times. Due to the randomness of the MC-Dropout, the predictions slightly change, and we can sample the model outputs' distribution. A spatial separation, i.e., clustering, of the inferred distributions is needed to study the statistical properties of each inferred distribution individually. Thus, we applied two clustering algorithms to this task, \emph{BGM\footnote{sklearn.mixture.BayesianGaussianMixture}}~\cite{Blei2006} instead of GMM as in \cite{Heidecker2021a} and \emph{Agglomerative Hierarchical Clustering\footnote{sklearn.cluster.AgglomerativeClustering} (AGG)}~\cite{Hastie2009} from the well-known python library sklearn~\cite{Pedregosa2011}. The disadvantage of GMM is that GMM adjusts a predefined number of components comparable to AGG. At the same time, BGM automatically infers the effective number of components from the data, and we only need to specify an upper limit. The number of components was calculated by dividing the number of predicted instances in the image by the number of repetitions $n$. Fig.~\ref{fig:cluster} describes the conceptual idea of our approach. For both clustering methods, we used the list of sampled bounding boxes $bbox=(x_{1}, y_{1}, x_{2}, y_{2})$ as input features, which holds the information for the spatial position, as well as the size of the predicted bounding box. When using BGM, another problem appeared, BGM tended to group high-density regions close to each other. On the one hand, this grouping is positive when one instance is split by another instance, forming multiple regions of high density. However, it also has a negative effect. For example, when two people stand close to each other, further away from the camera, and get grouped. For this reason, we reprocessed clusters of more than $150$ instances with BGM and thus broke the cluster into several clusters depending on the number of instances. The used threshold of $150$ showed good performance for the dataset we used. After clustering the bounding boxes, we sorted the lists of classes and masks based on their corresponding bounding box cluster. Finally, each cluster or instance consists of a \emph{bounding box list}, a \emph{class score list}, and a \emph{mask list}. We tried to include other features in the clustering, e.g., mask, but this did not yield better results than just using the bounding boxes.

\section{Evaluation}\label{sec:eval}
    The evaluation of the implemented methods provides information on how the changes affect the modeling of epistemic uncertainty. The model performance of the different variants is shown in Table~\ref{tab:model_perform}. Each model variant was trained on the \emph{COCO}~\cite{Lin2015} train dataset with a few images excluded for calibration. The calibration was performed on our sub-split of the COCO train dataset with around 4000 randomly selected images, and the \emph{COCO} validation dataset was used for evaluation. We trained all models for 20 epochs with all 80 available classes of the \emph{COCO} dataset plus the background class. Table~\ref{tab:model_perform} contains beside our tested model variants no.~3 to 12, the original model in row no.~1, and the results of the calibrated original model in row no.~2. As a performance metric, we used mean average precision (mAP) and the COCO-Evaluator from the COCO-API~\cite{Lin2015} for the bounding boxes and masks. Comparing the results of the two cluster algorithms used, BGM and AGG, it is evident that BGM (no.~5 to no.~8) provides slightly better results than AGG, independent of focal loss or calibration. The influence of focal loss, calibration, and MC-Dropout will be discussed below.
    
    Furthermore, we have detected that some clusters contain bounding boxes but do not have a corresponding mask, called zero masks, which affects the evaluation of the mask uncertainty. We identified two issues causing this: internal dependencies of Mask-RCNN (Section~\ref{subsec:internal}) or MC-Dropout (Section~\ref{subsec:mc_drop}).

    \subsection{Internal Dependencies}\label{subsec:internal}
        Internal dependencies of the classification head and the mask head could be one explanation for the zero masks. The mask head has an output shape of $k*m^2$ for each instance, where $k$ is the number of classes, and $m*m$ is the instance size. However, the final output of the mask head is a single mask out of $k$ selected by the classification head. If the wrong mask is selected due to the wrong class, it is possible that the mask head prediction is bad. We added focal loss and calibrated the model for more reliable class predictions (see Section \ref{subsec:focal_cal}). In addition to the selection of the appropriate mask, two different threshold values have a strong influence on the appearance of the mask. The mask output of Mask-RCNN is a binary mask created using a threshold of $0.5$ in the postprocessing step. The second threshold is applied after our clustering to the mean mask in a cluster to obtain a binary mask. This threshold is also set to $0.5$. If the masks in a cluster are not on top of each other (See Fig.~\ref{fig:heatmap}) and are rather distributed in the area and maybe also relatively poor, the threshold value causes the binary mask of the cluster to be very poor or even zero as non-existent.
    
    \subsection{Focal Loss and Calibration}\label{subsec:focal_cal}
        As mentioned in Section \ref{sec:mask_rcnn_plus}, we changed the loss function of Mask-RCNN by replacing the cross entropy loss with \emph{focal loss}~\cite{Lin2017} to compensate for the class imbalance. 
        \begin{equation}
            \mathrm{FL}(p_{t})=-\alpha_{t}(1-p_{t})^{\gamma}\mathrm{log}(p_{t}),
        \end{equation}\label{eq:fl}
        thereby $p_{t}$ describes the probability of the ground truth class, while $\alpha_{t}$ introduces weights to give small classes a higher weight than dominant classes. Because $\alpha_{t}$ does not differentiate between easy and hard samples, a tunable factor $\gamma$ is used to focus on hard negative samples during training \cite{Lin2017}. Comparing the results from no.~5 and no.~7 (without calibration) in Table~\ref{tab:model_perform} shows that the use of focal loss negatively affects the model.
    
\begin{figure}[t]
            \centering
            \begin{subfigure}[h]{0.48\textwidth}
                 \centering
                 \includegraphics[width=\textwidth]{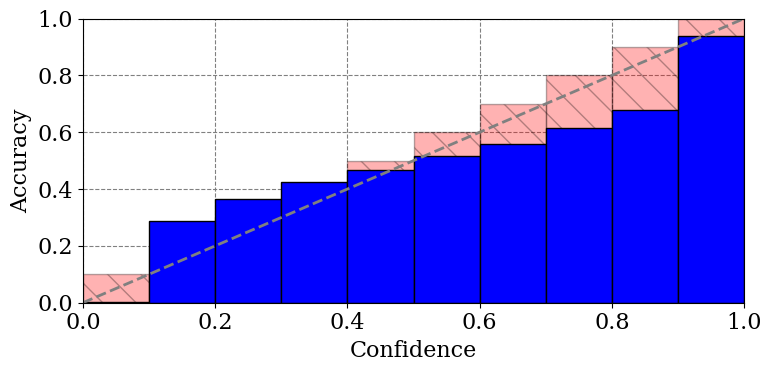}
             \end{subfigure}
             \vfill
             \begin{subfigure}[h]{0.48\textwidth}
                 \centering
                 \includegraphics[width=\textwidth]{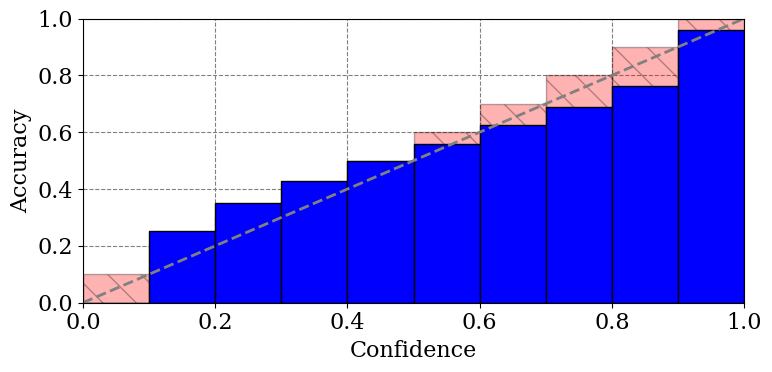}
             \end{subfigure}
            \caption{Reliability diagram of the classification head logits before (top) and after (bottom) model calibration.}
            \label{fig:cal}
            \vspace{-0.1cm}
        \end{figure}

        Mask-RCNN is, by default, uncalibrated, which means the prediction confidence score does not reflect the actual quality of the prediction. The confidence value should match the predicted class's actual correctness probability, i.e., accuracy, for an ideally calibrated model. However, there is often a gap between the ideal and the predicted confidence score, as shown in Fig.~\ref{fig:cal} on the top. 
        
        In Mask-RCNN, the predictions confidence scores $p_{i}$ with $i \in [1, \ldots, N]$ for each input $x_{i}$ are calculated by a softmax layer.
        \begin{equation}\label{eq:softmax}
            p_{i} = \underset{k}{\mathrm{max}}\ \sigma_{SM} ({z}_{i})^{(k)},\ 
            \sigma_{SM} ({z}_{i})^{(k)}=\frac{e^{z_{i}^{(k)}}}{\sum_{j=0}^{K}e^{{z_{i}^{(j)}}}}\text{~,}
        \end{equation}
        whereby $\sigma$ denotes the softmax function, the class logits are represented by $z_{i}$, $k$ defines the index of the input vector, and $K$ is the number of classes. As mentioned before, we applied \emph{Temperature Scaling}~\cite{Guo2017} for the classification head to improve the performance. \emph{Temperature Scaling} is a parametric approach where a single parameter $T$ in equ.~\ref{eq:temp}, called \emph{temperature}, is optimized to the negative log-likelihood on the validation set and used to scale all the classes without changing the maximum of the softmax function.
        \begin{equation}\label{eq:temp}
            \hat{p_{i}} = \underset{k}{\mathrm{max}}\, \sigma_{SM}(\mathbf{z}_{i}/T)^{(k)}
        \end{equation}
        The calibration to the Mask-RCNN model reduced the gap difference between the ideal and predicted confidence score, as shown in the bottom of Fig.~\ref{fig:cal}. The Maximum Calibration Error decreased from $0.176$ to $0.109$, and the Average Calibration Error from $0.083$ to $0.055$. If we compare the uncalibrated (no.~5 and no.~7) and calibrated cases (no.~6 and no.~8) of Table~\ref{tab:model_perform}, the calibration slightly improves the model performance. Overall, for a model with a dropout rate of $0.5$, we have the best results without focal loss but with calibration.

    \begin{figure}[t]
        \centering
        \begin{subfigure}[h]{0.45\textwidth}
            \centering
            \includegraphics[width=\textwidth]{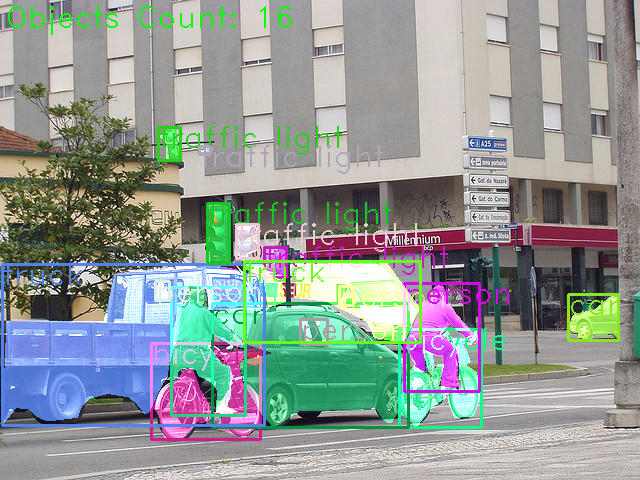}
            \caption{Ground truth.}
        \end{subfigure}
        \vfill
        \vspace{0.1cm}
        \begin{subfigure}[h]{0.45\textwidth}
            \centering
            \includegraphics[width=\textwidth]{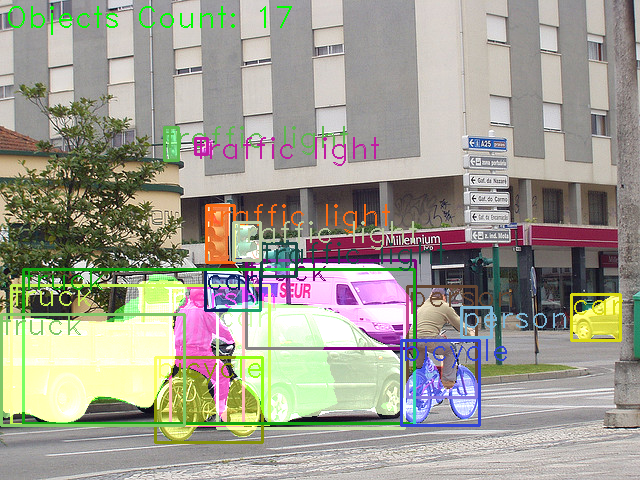}
            \caption{Model prediction.}
        \end{subfigure}
        \caption{Image example of a street scene with the ground truth (top) and model predictions (bottom). Each instance has a bounding box and mask. The mask color is random and has no relation with the instance class.}
        \label{fig:gt_pred}
        \vspace{-0.1cm}
    \end{figure}

    \begin{figure*}[t]
        \centering
        \begin{subfigure}[h]{0.45\textwidth}
            \centering
            \includegraphics[width=\textwidth]{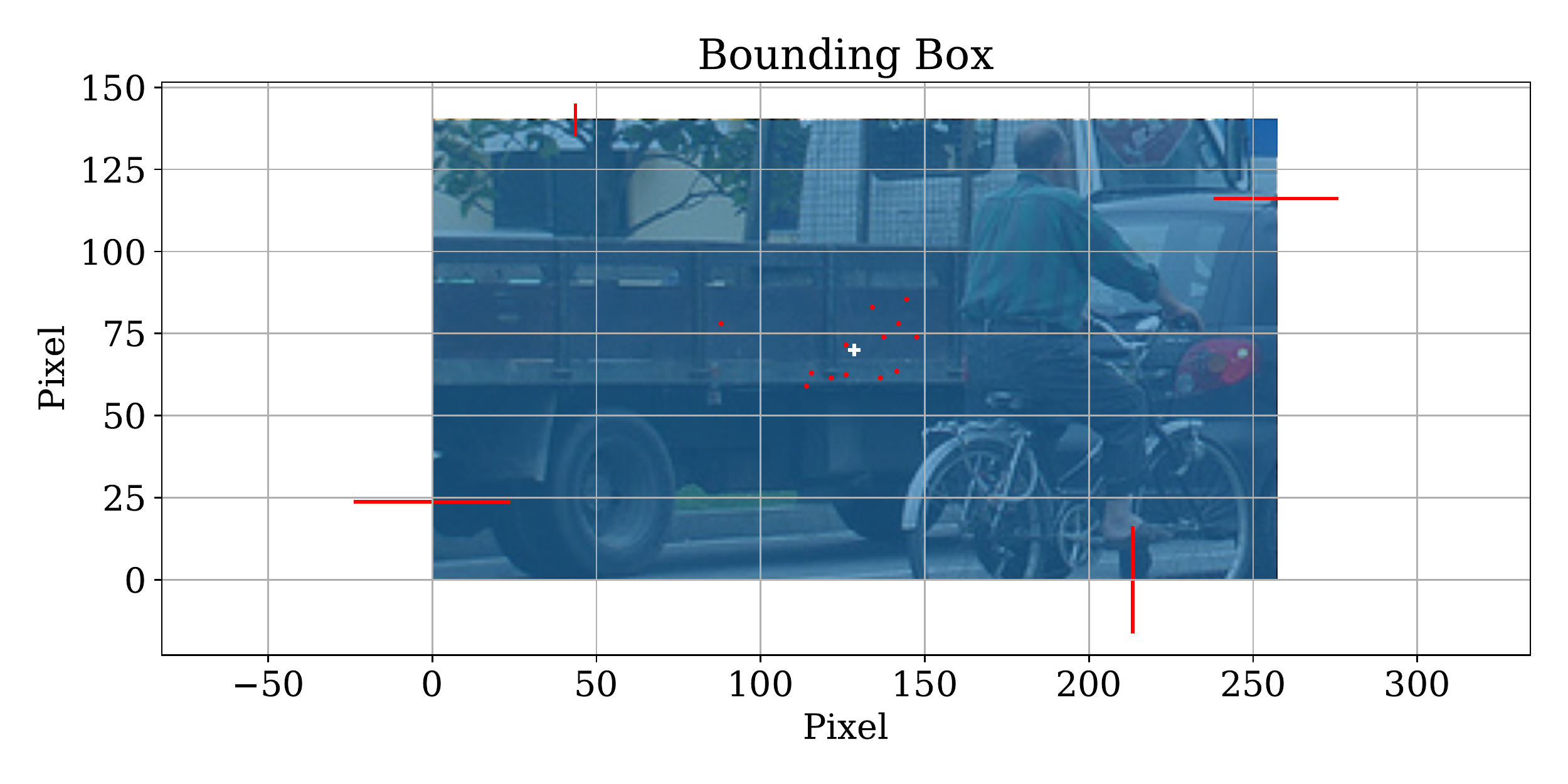}
        \end{subfigure}
        \hfill
        \begin{subfigure}[h]{0.45\textwidth}
            \centering
            \includegraphics[width=\textwidth]{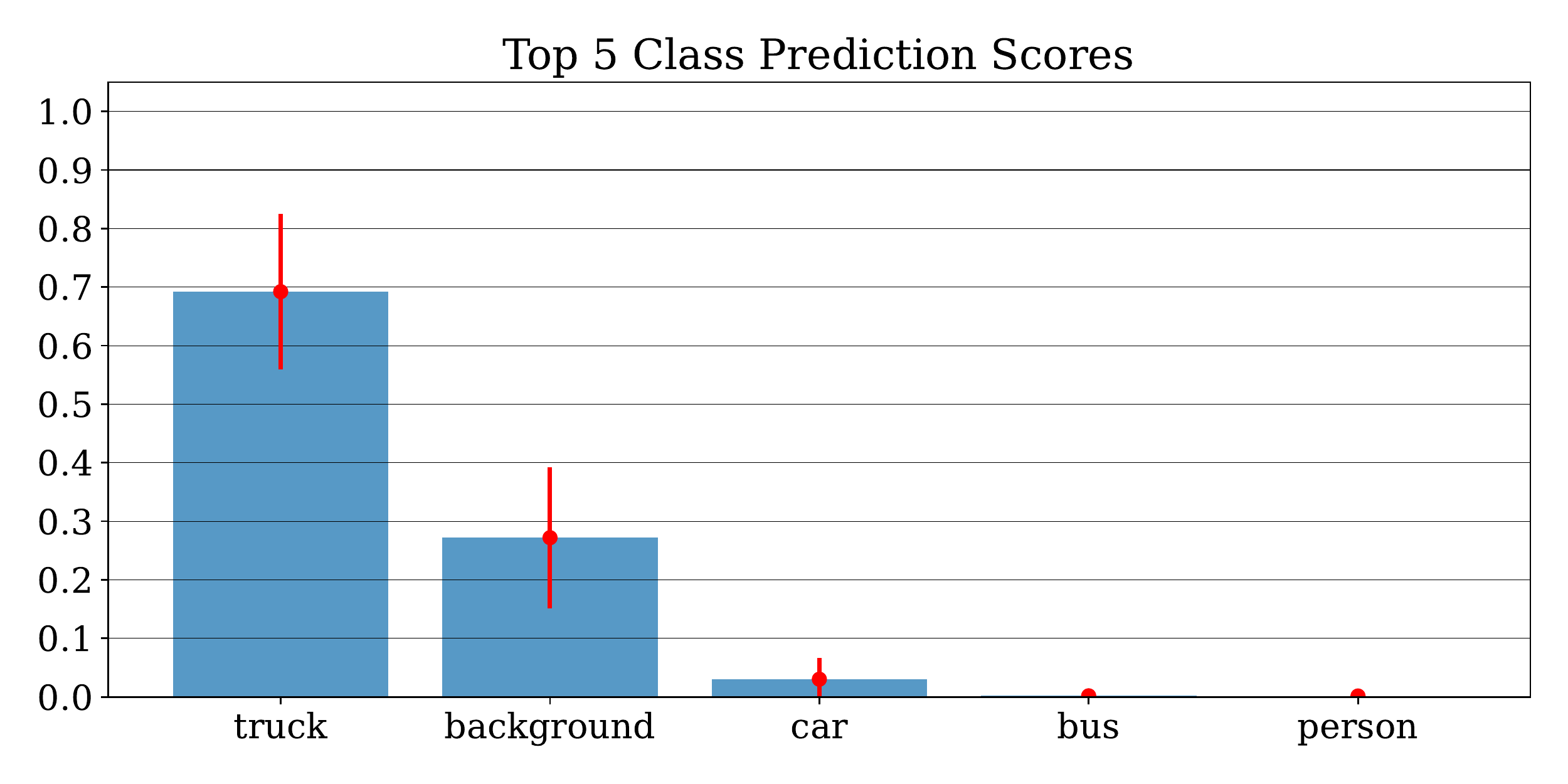}
        \end{subfigure}
        \vfill
        \begin{subfigure}[h]{0.45\textwidth}
            \centering
            \includegraphics[width=\textwidth]{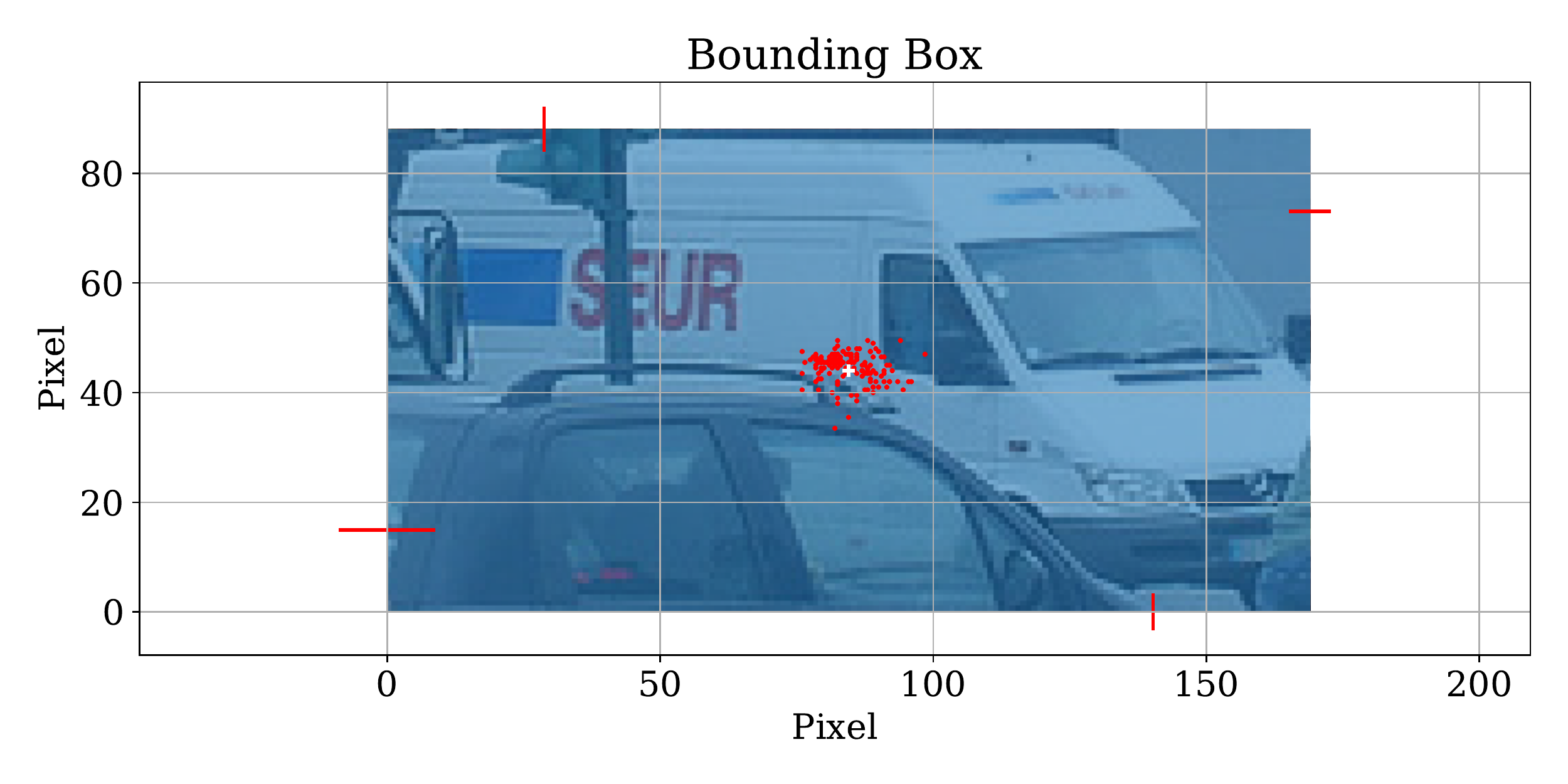}
        \end{subfigure}
        \hfill
        \begin{subfigure}[h]{0.45\textwidth}
            \centering
            \includegraphics[width=\textwidth]{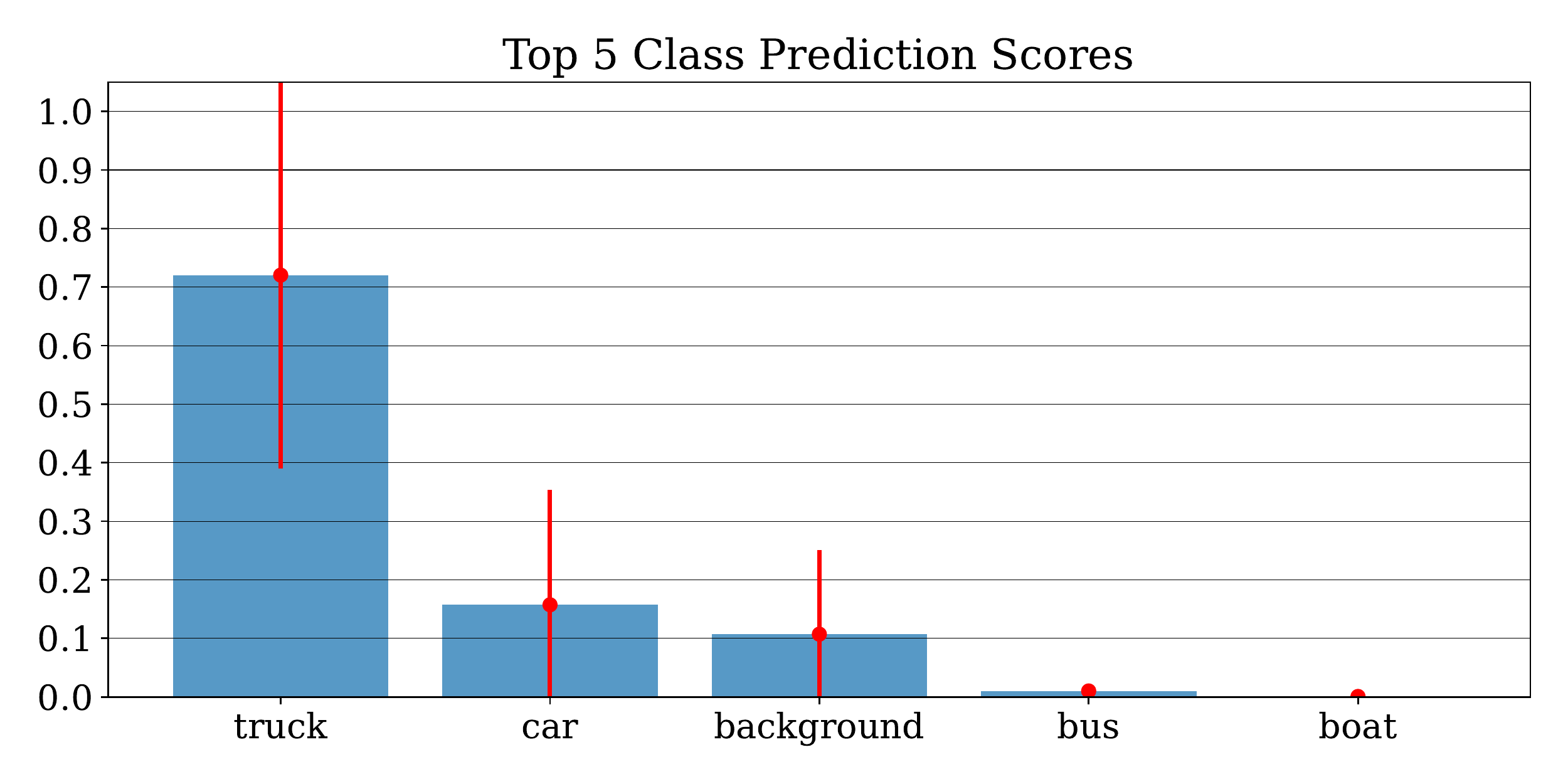}
        \end{subfigure}
        \vfill
        \begin{subfigure}[h]{0.45\textwidth}
            \centering
            \includegraphics[width=\textwidth]{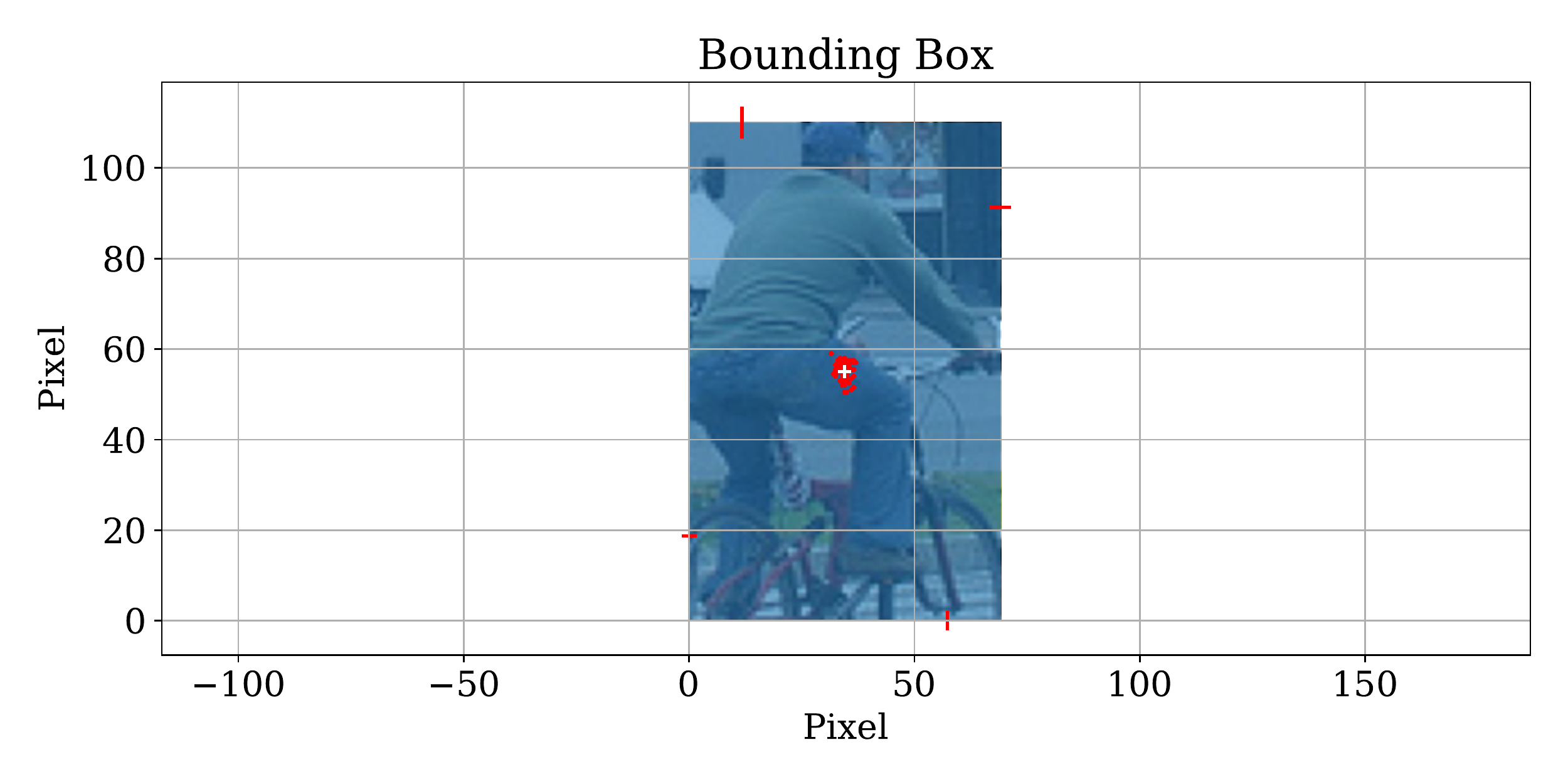}
        \end{subfigure}
        \hfill
        \begin{subfigure}[h]{0.45\textwidth}
            \centering
            \includegraphics[width=\textwidth]{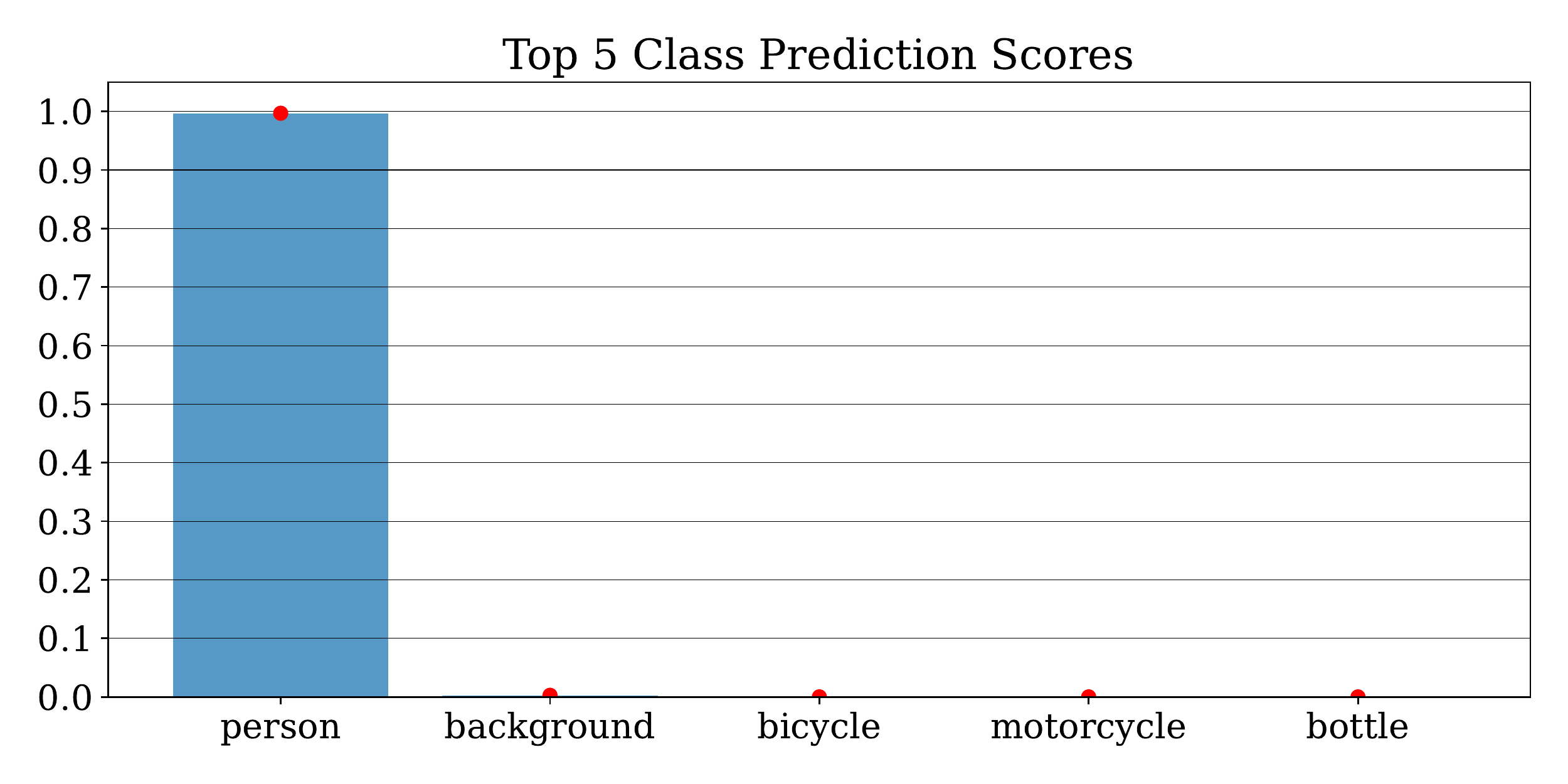}
        \end{subfigure}
        \caption{Uncertainty visualization for bounding box (left) and class (right), of two instances of Fig.~\ref{fig:gt_pred}.}
        \label{fig:uncertainty_bbox_class}
    \end{figure*}

    \subsection{MC-Dropout}\label{subsec:mc_drop}
        Another explanation for the zero masks phenomenon could be the MC-Dropout layers added to the mask head, as shown in Fig.~\ref{fig:mask_rcnn_plus}. Since the dropout layer reduces the averaged sum activation value of the previous layer, it could be the case where the dropout rate value is too high that it dampens the neurons' activation, resulting in zero masks. In \cite{Gal2017}, different methods for deciding on the best dropout rate are discussed, which controls the percentage of the inactive (dropped) neurons in a specific layer. However, even in convolutional neural networks, dropout was usually implemented in the fully connected layers, not in the convolutional layers. Park et al.~\cite{Park2017} conducted a study on the effect of MC-Dropout on convolutional neural networks. To assess the dropout, we observe the average activation of the neurons in the feature detectors after each dropout layer, as proposed by \cite{Park2017}. \cite{Park2017} continues, it is natural to have a relatively low average activation for a specific layer that lies deeper in the model. The shallower layers at the beginning tend to have higher average activation of neurons than their deeper counterparts. This means that we should consider different dropout rates depending on the relative location of the layer in the model. For comparison, we selected a dropout rate of $0.2$, $0.5$, and a combination of both, called $mix$. With our results in Table~\ref{tab:model_perform}, we observe that with a dropout rate of $0.2$ (no.~3), we archive a better result than  with $0.5$ (no.~5), which is understandable due to less dropped neurons. The results of the $mix$ dropout rate (no.~4) are nearly identical to the $0.5$ dropout rate, where we would have expected to come close to the result of the $0.2$ dropout rate.

\section{Visualization}\label{sec:visual}
    The following four sections present different visualizations that visualize the encountered uncertainty. To illustrate the uncertainty of the bounding box head, the classification head, and the mask head, we chose the example shown in Fig. \ref{fig:gt_pred}, with the ground truth (top), along with the prediction of the model (bottom).
    
    \subsection{Box Uncertainty Visualization} \label{sub:bbox}
        To visualize the uncertainty in the bounding box head, we consider the parameters that characterize the distribution of the sampled boxes. These parameters are the mean and the standard deviation of the boxes representing this cluster. The mean is considered the box representing the whole cluster in a simplified manner. The standard deviation describes the variability in the box size that spans two dimensions $[x, y]$, as well as the variations in the sampled boxes' locations across the cluster. Fig.~\ref{fig:uncertainty_bbox_class} shows the mean bounding box of three examples on the left. The white cross denotes the bounding box center. The red dots represent each bounding box's center point of the whole instance cluster. The red lines describe the standard deviation of each of the four bounding box edges. 
        
        In the example of the truck (top left) in Fig.~\ref{fig:uncertainty_bbox_class}, we see that the standard deviation in each direction is significant, most likely caused by the other instance in the foreground. Both other examples are performing better, but they are not perfect. The truck (middle) shows a higher standard deviation to the left, which is understandable because the model has to separate both trucks from each other. In addition, the center points (red dots) are slightly distributed horizontally, which supports this argument.

        \begin{figure}[t]
            \centering
            \begin{subfigure}[h]{0.45\textwidth}
                \centering
                \includegraphics[width=\textwidth]{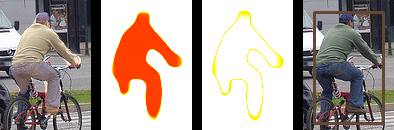}
                \caption{}\label{fig:heatmap_a}
            \end{subfigure}
            \vfill
            \vspace{0.1cm}
            \begin{subfigure}[h]{0.45\textwidth}
                \centering
                \includegraphics[width=\textwidth, trim={0, 0, 585, 0}, clip]{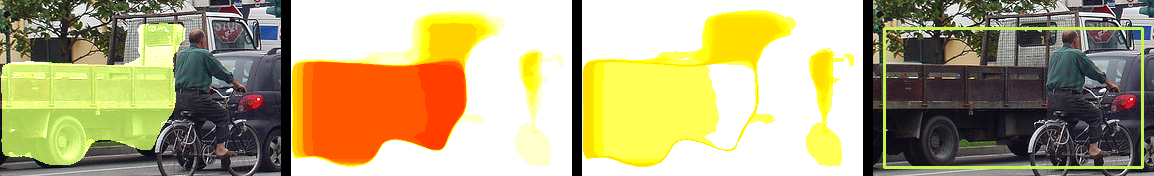}
            \end{subfigure}
            \begin{subfigure}[h]{0.45\textwidth}
                \centering
                \includegraphics[width=\textwidth, trim={585, 0, 0, 0}, clip]{images/obj_5_heatmap.png}
                \caption{}\label{fig:heatmap_b}
            \end{subfigure}
            \caption{Instance mask uncertainty. Each of the two examples contains four images. The first and last present the instance binary mask or box overlay, the second presents the mean mask, and the third describes the standard deviation of the mask cluster.}
            \label{fig:heatmap}
            \vspace{-0.1cm}
        \end{figure}
    
    \subsection{Classification Uncertainty Visualization} \label{sub:class}
        The predicted class of the instance is simply the highest-scored class in the classification head. We also included the background class because it is dominant and always present under the top 5 class scores in our examples. The right column of Fig.~\ref{fig:uncertainty_bbox_class} shows the class score mean and standard deviation. Each segment represents a class denoted on the x-axis. The red dot in the middle of each segment represents the mean score of this particular class, while the whiskers represent a single standard deviation.
        
        In the right column of Fig.~\ref{fig:uncertainty_bbox_class}, the class probability is plotted for the three examples from the left. We can see that the correct class has the highest mean score, but the standard deviation is also high for some other classes. This indicates that within the cluster, the model sometimes picks the wrong class; therefore, the model is not always sure about the classification, especially if the standard deviation is high, as in the middle example.

        \begin{figure}[htp]
            \centering
            \begin{subfigure}[h]{0.49\textwidth}
                 \centering
                 \includegraphics[width=\textwidth]{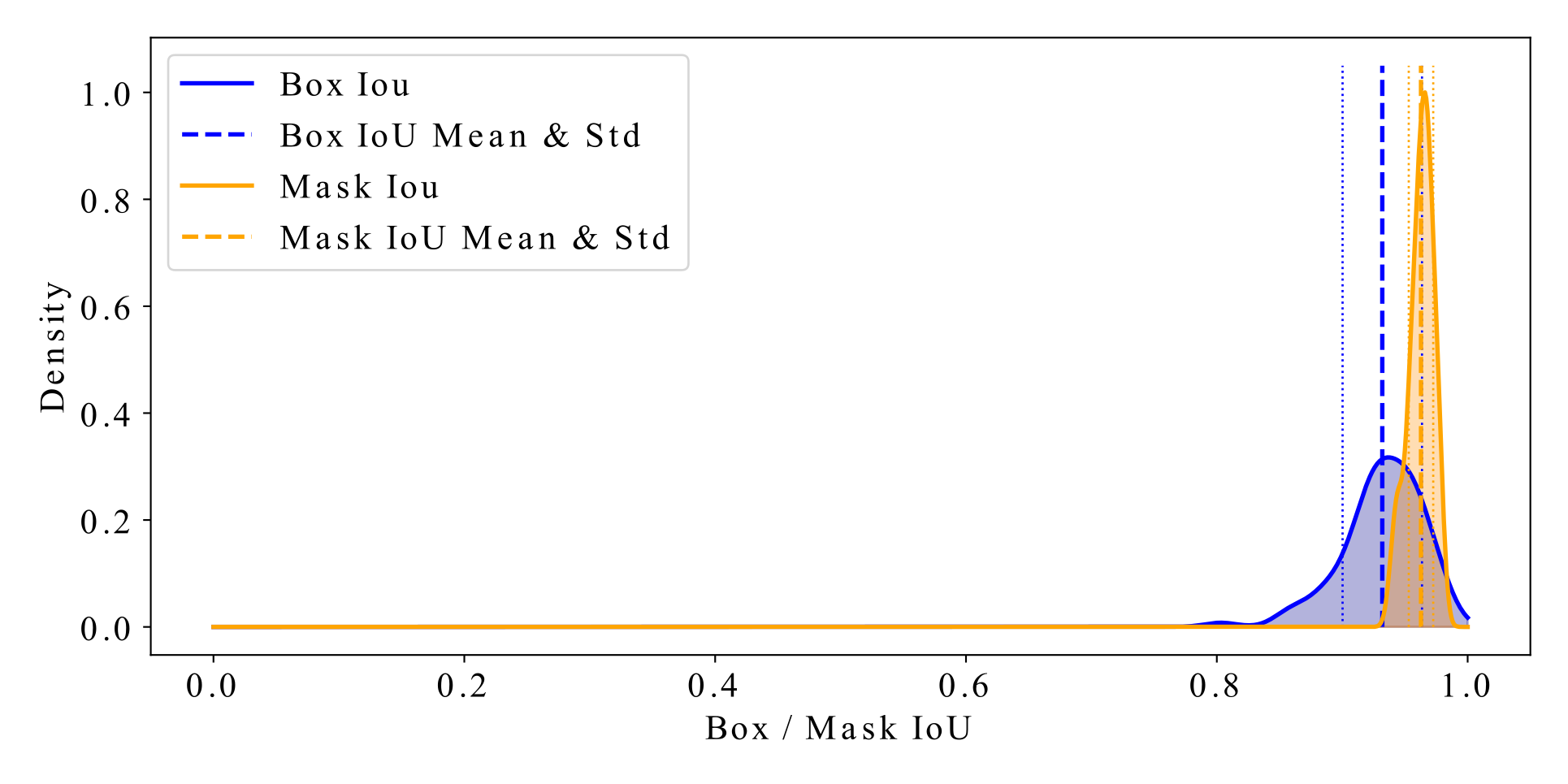}
                 \caption{KDE of Fig.~\ref{fig:heatmap_a}.}\label{fig:kde_a}
             \end{subfigure}
             \vfill
             \begin{subfigure}[h]{0.49\textwidth}
                 \centering
                 \includegraphics[width=\textwidth]{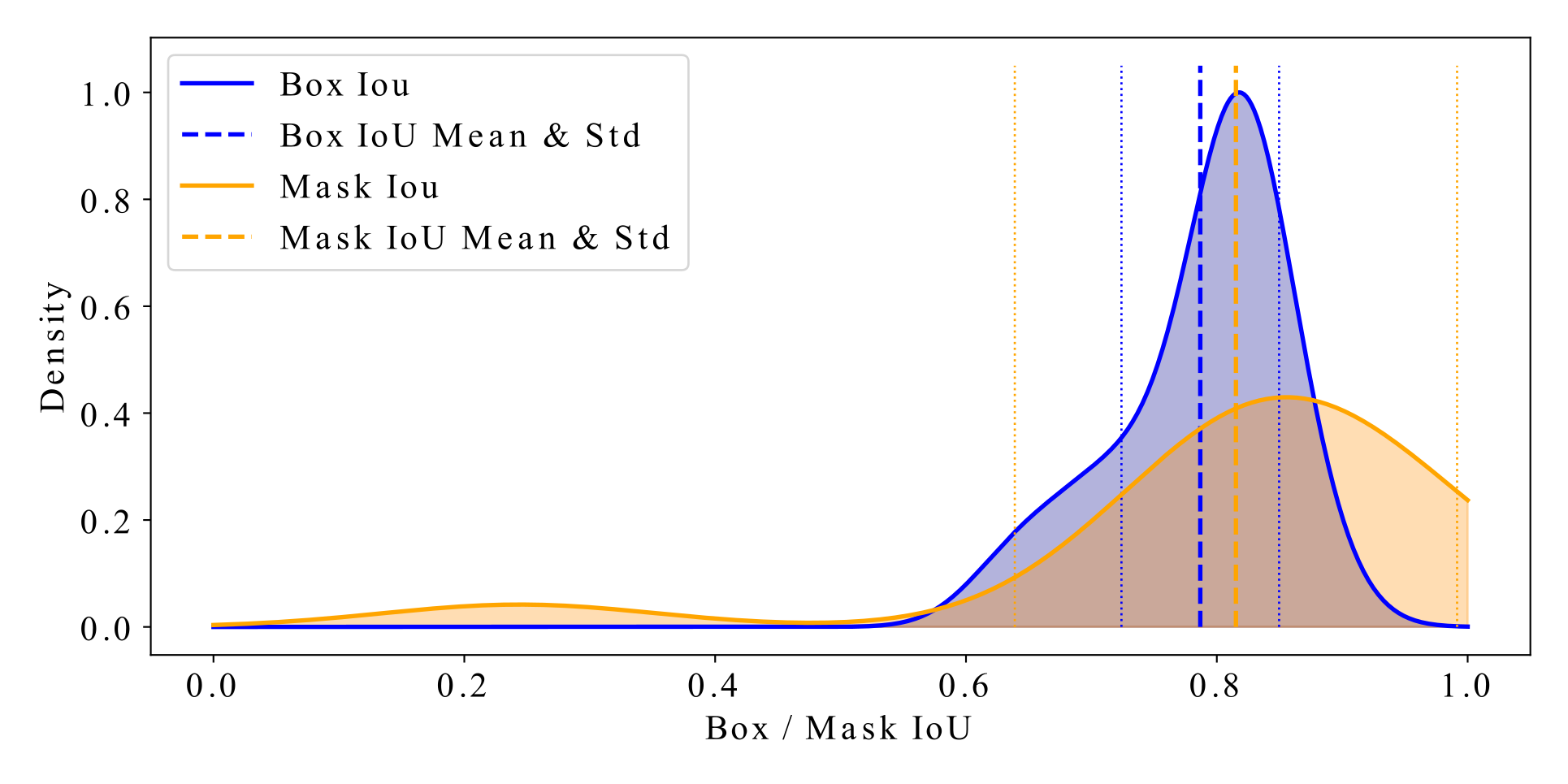}
                 \caption{KDE of Fig.~\ref{fig:heatmap_b}.}\label{fig:kde_b}
             \end{subfigure}
            \caption{Kernel density estimation plot for predicted boxes IoU (blue) and predicted masks IoU (orange). The big dashed line represents the mean, and the fine dashed line on both sides visualizes a single standard deviation.}
            \label{fig:kde}
        \end{figure}

    \subsection{Mask Uncertainty Visualization}\label{sub:heatmap}
        To represent the semantic masks' uncertainty, each pixel uncertainty is represented in a heat map, Fig.~\ref{fig:heatmap}. The redder the color, the higher the certainty of this pixel. The model is neither certain nor uncertain in the white areas of the heat map, as we have no knowledge of these areas. 
        
        This visualization contains all the masks that the model has predicted within the instance cluster. We can see how well the mask head recognizes the instance. For example, in the case of the person riding a bike, the mask (Fig.~\ref{fig:heatmap_a}: first and second image) covers the rider very well. The mask standard deviation is very low and high only along the contour of the rider (Fig.~\ref{fig:heatmap_a}: third image). The truck represents the opposite, the mask is not covering the whole instance (Fig.~\ref{fig:heatmap_b}: first image), and the standard deviation (Fig.~\ref{fig:heatmap_b}: third image) is not just along the instance contour.
            
    \subsection{Kernel Density Estimation Plots} \label{sub:kde}
        The IoU is widely used in evaluating object detection models. It describes the quality of the predictions, e.g., semantic masks and bounding boxes. Conventionally, calculating the IoU of a predicted mask or box requires the ground truth label of the object as a reference to assess prediction quality. In this work, we used the bounding box mean to reference each cluster member for calculating the IoU. Therefore, the IoU value can be interpreted as the distance between each prediction and the mean prediction across all prediction samples. In Fig.~\ref{fig:kde}, we plot the kernel density of the IoU values, transforming the uncertainty in another space. Using kernel density estimation (KDE) allows us to compare different prediction domains, e.g., semantic masks and bounding boxes.

        In Fig.~\ref{fig:kde_a}, the distribution of the mask is very narrow, which is an indication that all masks are more or less identical. The same observation we made in Fig.~\ref{fig:heatmap_a} by evaluating the mask of the rider. The box IoU is also well and confirms the standard deviation of the bounding box edges in Fig.~\ref{fig:uncertainty_bbox_class} (bottom left). Both distributions in Fig.~\ref{fig:kde_a} are not identical, but they are similar to each other than both in Fig.~\ref{fig:kde_b}. The distributions in Fig.~\ref{fig:kde_b} reveal the same behavior (high standard deviation) we have already discussed together with the mask in Fig.~\ref{fig:heatmap_b} and the bonding box edges in Fig.~\ref{fig:uncertainty_bbox_class} (top left).

\section{Conclusion}\label{sec:conclusion}
    In this article, we have presented our modified architecture of Mask-RCNN to model epistemic uncertainty. We added MC-Dropout to each head of the model and the RPN. Through multiple repetitions of the MC-Dropout extended Mask-RCNN Model and the use of clustering, we showed that the model uncertainty of each object class, bounding box, and instance mask could be described. In terms of clustering, we examined BGM and AGG, with BGM performing better than AGG. By applying different dropout rates, we discovered that the model performance slightly decreases with increasing dropout rates. Besides, we also added focal loss to the classification head and calibrated the model. We evaluated the model performance, showing that a well-calibrated model outperforms focal loss slightly. To illustrate the modeled uncertainty, we created corresponding visualizations for each of the three considered model outputs bounding box, class score, and instance mask. 
    


    \section*{Acknowledgment}
    
    This work results from the project KI Data Tooling (19A20001O) funded by the German Federal Ministry for Economic Affairs and Climate Action (BMWK).

    \bibliographystyle{IEEEtran}
    \bibliography{bibliography}

\end{document}